# Small steps and giant leaps: Minimal Newton solvers for Deep Learning

João F. Henriques    Sebastien Ehrhardt    Samuel Albanie    Andrea Vedaldi
Visual Geometry Group, University of Oxford
{joao,hyenal,albanie,vedaldi}@robots.ox.ac.uk

## Abstract

We propose a fast second-order method that can be used as a drop-in replacement for current deep learning solvers. Compared to stochastic gradient descent (SGD), it only requires two additional forward-mode automatic differentiation operations per iteration, which has a computational cost comparable to two standard forward passes and is easy to implement. Our method addresses long-standing issues with current second-order solvers, which invert an approximate Hessian matrix every iteration exactly or by conjugate-gradient methods, a procedure that is both costly and sensitive to noise. Instead, we propose to keep a single estimate of the gradient projected by the inverse Hessian matrix, and update it once per iteration. This estimate has the same size and is similar to the momentum variable that is commonly used in SGD. No estimate of the Hessian is maintained. We first validate our method, called CURVEBALL, on small problems with known closed-form solutions (noisy Rosenbrock function and degenerate 2-layer linear networks), where current deep learning solvers seem to struggle. We then train several large models on CIFAR and ImageNet, including ResNet and VGG-f networks, where we demonstrate faster convergence with no hyperparameter tuning. Code is available.

## 1 Introduction

Stochastic Gradient Descent (SGD) and back-propagation [9] are the algorithmic backbone of current deep network training. The success of deep learning demonstrates the power of this combination, which has been successfully applied on various tasks with large datasets and very deep networks [7].

Yet, while SGD has many advantages, speed of convergence (in terms of number of iterations) is not necessarily one of them. While individual SGD iterations are very quick to compute and lead to rapid progress at the beginning of the optimisation, soon the latter reaches a slower phase where further improvements are achieved slowly. This can be attributed to entering regions of the parameter space where the objective function is poorly scaled. In such cases, rapid progress would require vastly different step sizes for different directions in parameter space, which SGD cannot deliver.

Second-order methods, such as Newton's method and its variants, eliminate this issue by rescaling the gradient according to the local curvature of the objective function. For a scalar loss in $\mathbb{R}$, this rescaling takes the form $H^{-1}J$ where $H$ is the Hessian matrix (second-order derivatives) or an approximation of the local curvature in the objective space, and $J$ is the gradient of the objective. They can in fact achieve local scale-invariance, and make provably better progress in the regions where gradient descent stalls [24]. While they are unmatched in other domains, several problems have impeded their application to deep models. First, it is impractical to invert or even store the Hessian matrix, since it grows quadratically with the number of parameters, and there are typically millions of them. Second, any Hessian estimate is necessarily noisy and ill-conditioned due to stochastic sampling, to which classic inversion procedures such as conjugate-gradient are not robust.

In this paper, we propose a new algorithm that can overcome these difficulties and make second order optimisation practical for deep learning. We show in particular how to avoid the storage of any estimate of the Hessian matrix or its inverse. Instead, we treat the computation of the Newton update, $H^{-1}J$, as solving a linear system that itself can be solved via gradient descent. The cost of solving this system is amortized over time by interleaving its steps with the parameter update steps. Moreover, the choice of gradient descent makes it robust to noise, unlike conjugate-gradient methods. Our proposed method adds little overhead, since a Hessian-vector product can be implemented for modern networks with just two steps of automatic differentiation. Interestingly, we show that our method is equivalent to momentum SGD (also known as the heavy-ball method) with a single additional term, accounting for curvature. For this reason we named our method CURVEBALL. Unlike other proposals, the total memory footprint is as small as that of momentum SGD.

This paper is structured as follows. We introduce relevant technical background in sec. 2, and present our method in sec. 3. We evaluate our method and show experimental results in sec. 4. Related work is discussed in sec. 5. Finally we summarise our findings in sec. 6.

## 2 Background

In order to make the description of our method self-contained, we succinctly summarise a few standard concepts in optimisation. Our goal is to find the optimal parameters of a model (*e.g.* a neural network) $\phi : \mathbb{R}^p \to \mathbb{R}^o$, with $p$ parameters $w \in \mathbb{R}^p$ and $o$ outputs (the notation does not show the dependency on the training data, which is subsumed in $\phi$ for compactness). The quality of the outputs is evaluated by a loss function $L : \mathbb{R}^o \to \mathbb{R}$, so finding $w$ is reduced to the optimisation problem: [1]

$$w^* = \arg\min_w L(\phi(w)) = \arg\min_w f(w). \tag{1}$$

Probably the simplest algorithm to optimise eq. 1 is gradient descent (GD). GD updates the parameters using the iteration $w \leftarrow w - \alpha J(w)$, where $\alpha > 0$ is the learning rate and $J(w) \in \mathbb{R}^p$ is the gradient (or Jacobian) of the objective function $f$ with respect to the parameters $w$. A useful variant is to augment GD with a *momentum* variable $z$, which can be interpreted as a decaying average of past gradients:

$$z \leftarrow \rho z - J(w) \tag{2}$$
$$w \leftarrow w + \alpha z \tag{3}$$

Momentum GD, as given by eq. 2-3, can be shown to have faster convergence than GD, remaining stable under higher learning rates, and exhibits somewhat better resistance to poor scaling of the objective function [6]. One important aspect is that these advantages cost almost no additional computation and only a modest additional memory, which explains why it is widely used in practice.

In neural networks, GD is usually replaced by its stochastic version (SGD), where at each iteration one computes the gradient not of the model $f = L(\phi(w))$, but of the model $f_t = L_t(\phi_t(w))$ assessed on a small batch of samples, drawn at random from the training set.

### 2.1 Second-order optimisation

As mentioned in section 1, the Newton method is similar to GD, but steers the gradient by the inverse Hessian matrix, computing $H^{-1}J$ as a descent direction. However, inverting the Hessian may be numerically unstable or the inverse may not even exist. To address this issue, the Hessian is usually regularized, obtaining what is known as the *Levenberg* [14] method:

$$\triangle_w = -(H + \lambda I)^{-1} J, \tag{4}$$
$$w \leftarrow w + \alpha \triangle_w, \tag{5}$$

where $H \in \mathbb{R}^{p \times p}$, $J \in \mathbb{R}^p$, and $I \in \mathbb{R}^{p \times p}$ is the identity matrix. To avoid burdensome notation, we omit the $w$ argument in $H(w)$ and $J(w)$, but these quantities must be recomputed at each iteration. Intuitively, the effect of eq. 4 is to rescale the step appropriately for different directions — directions with high curvature require small steps, while directions with low curvature require large steps to make progress.

---

[1] We omit the optional regulariser term for brevity, but this does not materially change our derivations.

Note also that Levenberg's regularization loses the scale-invariance of the original Newton method, meaning that rescaling the function $f$ changes the scale of the gradient and hence the regularised descent direction chosen by the method. An alternative that alleviates this issue is *Levenberg-Marquardt*, which replaces $I$ in eq. 4 with $\text{diag}(H)$. For non-convex functions such as deep neural networks, these methods only converge to a local minimum when the Hessian is positive-semidefinite (PSD).

## 2.2 Automatic differentiation and back-propagation

In order to introduce fast computations involving the Hessian, we must take a short digression into how Jacobians are computed. The Jacobian of $L(\phi(w))$ (eq. 1) is generally computed as $J = J_\phi J_L$ where $J_\phi \in \mathbb{R}^{p \times o}$ and $J_L \in \mathbb{R}^{o \times 1}$ are the Jacobians of the model and loss, respectively. In practice, a Jacobian is never formed explicitly, but Jacobian-vector products $Jv$ are implemented with the back-propagation algorithm. We define

$$\overleftarrow{\text{AD}}(v) = Jv \tag{6}$$

as the *reverse-mode automatic differentiation* (RMAD) operation, commonly known as *back-propagation*. A perhaps lesser known alternative is *forward-mode automatic differentiation* (FMAD), which computes a vector-Jacobian product, from the other direction:

$$\overrightarrow{\text{AD}}(v') = v'J \tag{7}$$

This variant is less commonly-known in deep learning as RMAD is appropriate to compute the derivatives of a scalar-valued function, such as the learning objective, whereas FMAD is more appropriate for vector-valued functions of a scalar argument. However, we will show later that FMAD is relevant in calculations involving the Hessian.

The only difference between RMAD and FMAD is the direction of associativity of the multiplication: FMAD propagates gradients in the forward direction, while RMAD (or back-propagation) does it in the backward direction. For example, for the composition of functions $a \circ b \circ c$,

$$\overrightarrow{\text{AD}}_{a \circ b \circ c}(v) = ((vJ_a)J_b)J_c$$
$$\overleftarrow{\text{AD}}_{a \circ b \circ c}(v') = J_a(J_b(J_c v'))$$

Because of this, both operations have similar computational overhead, and can be implemented similarly. Note that, because the loss is a scalar function, the starting projection vector $v$ in back-propagation is a scalar and we set $v = 1$. For intermediate computations, however, it is generally a (vectorized) tensor of gradients.

## 2.3 Fast Hessian-vector products

Since the Hessian of learning objectives involving deep networks is not necessarily positive semi-definite (PSD), it is common to use the Gauss-Newton approximation [24, 12, 2]:

$$\hat{H} = J_\phi H_L J_\phi^T, \tag{8}$$

When $H_L$ is PSD, which is the case for all convex losses (*e.g.* logistic loss, $L^p$ distance), the resulting $\hat{H}$ is PSD by construction. Even though it is approximate, it is still desirable to use eq. 8 since it is guaranteed to be PSD and thus prevents second-order methods from being attracted to saddle-points [4].

For the method that we propose, and indeed for any method that implicitly inverts the Hessian (or its Gauss-Newton approximation), only computing Hessian-vector products $\hat{H}v$ is required. As such, eq. 8 takes a very convenient form:

$$\hat{H}v = J_\phi \left( H_L \left( J_\phi^T v \right) \right) \tag{9}$$
$$= \overleftarrow{\text{AD}}_\phi \left( H_L \left( \overrightarrow{\text{AD}}_\phi(v) \right) \right). \tag{10}$$

The cost of eq. 10 is thus equivalent to that of two back-propagation operations. This is similar to a classic result [17, 20], but written in terms of common automatic differentiation operations. The intermediate matrix-vector product $H_L u$ has negligible cost: for example, $H_L = 2I \Rightarrow H_L u = 2u$ for the squared-distance loss. Similarly, for the multinomial logistic loss we have $H_L = \text{diag}(p) - pp^T \Rightarrow H_L u = p \odot u - p(p^T u)$, where $p$ is the vector of predictions from a softmax layer and $\odot$ is the element-wise product. These products thus require only element-wise operations.

## 3 Method

This section presents the main contribution of our paper: a method that minimizes a second-order Taylor expansion of the objective (like the Newton variants from section 2.1), but at a much reduced computational and memory cost. The result of taking a step $\Delta_w$ away from a starting point $w$ can be modelled using a second-order Taylor expansion of the objective $f$:

$$f(w + \Delta_w) = \underbrace{f(w) + \Delta_w^T J(w) + \tfrac{1}{2} \Delta_w^T H(w) \Delta_w}_{\hat{f}(w, \Delta_w)} + \mathcal{O}(\|\Delta_w\|_2) \tag{11}$$

Most second-order methods seek the update $\Delta_w$ that minimizes $\hat{f}$, by ignoring the higher-order terms:

$$\triangle_w = \arg\min_z \hat{f}(z) = \arg\min_z \tfrac{1}{2} z^T H z + z^T J \tag{12}$$

In general, eq. 12 is solved by either explicit inversion $H^{-1}J$ [12, 2] or using the conjugate gradient method [11]. However, we suggest to simply optimise eq. 12 by gradient descent on $z$, with steps:

$$\triangle_z = J_{\hat{f}(z)} = Hz + J \tag{13}$$

When the linear system (Hessian and Jacobian) is fixed, this iteration always reaches the optimal solution. However, to amortize this cost over time we may forgo exact minimization, and instead interleave the optimisation of $z$ and $w$. In this case, the linear system continually changes as $H(w)$ and $J(w)$ are functions of the parameters $w$, and the role of the quantity $z$ is to track the solution of eq. 4 as the parameters change. By using GD for eq. 12, we ensure that the introduction of stochastic noise and small batches when optimising deep networks do not destabilize the inversion, since SGD is fairly robust under such conditions (unlike, say, conjugate-gradient). Putting the two updates together, we alternate between updating the linear system solution $z$ and the network parameters $w$:

$$\triangle_z = Hz + J \tag{14}$$
$$z \leftarrow \rho z - \beta \triangle_z \tag{15}$$
$$w \leftarrow w + \alpha z \tag{16}$$

where the gradient descent over $z$ has learning rate $\beta$, and we introduce a decay rate $\rho \leq 1$. The reason for $\rho$ (when $\rho < 1$) is so that $z$ gradually forgets old updates (since the linear system changes over time). The decay rate $\rho$ also admits an interesting interpretation as momentum, which also known as the heavy-ball method. By comparing eq. 14-16 to those for momentum SGD (eq. 2-3), we can see that they are almost equivalent. The only differences are that we add a curvature term $Hz$, and introduce the learning rate $\beta$ (which can be made redundant with $\beta = 1$). Due to the addition of curvature to the heavy-ball method, we decided to name our algorithm CURVEBALL.

**Implementation.** Using the fast Hessian-vector products from section 2.3, it is easy to implement eq. 14-16, including a regularization term $\lambda I$ (section 2.1). We can further improve eq. 14 by grouping the operations to minimize the number of automatic differentiation (back-propagation) steps:

$$\triangle_z = \left(J_\phi H_L J_\phi^T + \lambda I\right) z + J_\phi J_L \tag{17}$$
$$= J_\phi \left(H_L J_\phi^T z + J_L\right) + \lambda z \tag{18}$$

In this way, the total number of passes over the model is two: we compute $J_\phi v$ and $J_\phi^T v'$ products, implemented respectively as one RMAD (back-propagation) and one FMAD operation (section 2.2).

**Automatic $\rho$, $\alpha$ and $\beta$ hyper-parameters in closed form.** The method that we propose introduces a few hyper-parameters, which just like with SGD, would require tuning for different settings. Ideally, we would like to have no tuning at all. Fortunately, the quadratic minimization interpretation in eq. 12 allows us to draw on standard results in optimisation [24]. At any given step, the optimal $\rho$ and $\beta$ can be obtained by solving a simple $2 \times 2$ linear system [12]:

$$\begin{bmatrix} -\beta \\ \rho \end{bmatrix} = - \begin{bmatrix} \Delta_z^T \hat{H} \Delta_z & z^T \hat{H} \Delta_z \\ z^T \hat{H} \Delta_z & z^T \hat{H} z \end{bmatrix}^{-1} \begin{bmatrix} J^T \Delta_z \\ J^T z \end{bmatrix} \tag{19}$$

Note that, in calculating the proposed update (eq. 17), the quantities $\Delta_z$, $J_\phi^T z$ and $J_L$ have already been computed and can now be reused. Together with the fact that $\hat{H} = J_\phi H_L J_\phi^T$, this means that the elements of the above $2 \times 2$ matrix can be computed with only one additional forward pass. As for the learning rate $\alpha$, the optimal fitting suggests that it should be simply set to 1 [12].

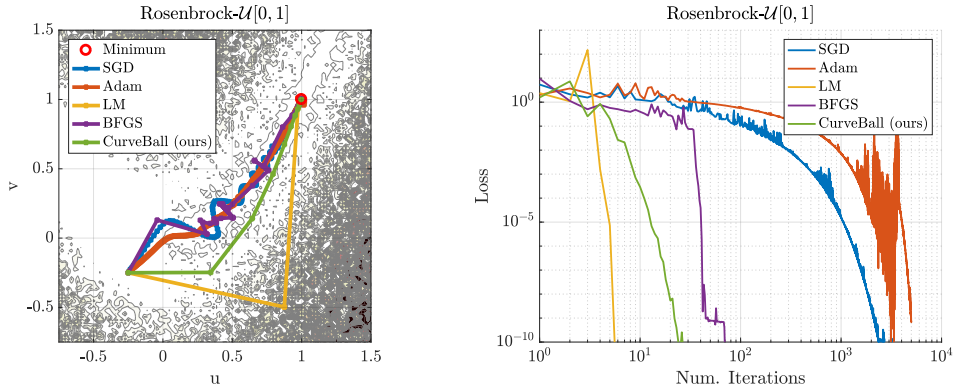

Figure 1: **Problems with known solutions**. Left: Trajectories on the Stochastic Rosenbrock function for different solvers (darker shaded regions denote higher function values). Right: evolution of the loss per iterations for the plotted trajectories.

**Automatic $\lambda$ hyper-parameter rescaling.** The regularization term $\lambda I$ (eq. 4) can be interpreted as a trust-region [24]. When the second-order approximation holds well, $\lambda$ can be small, corresponding to an unregularized Hessian and a large trust-region. Conversely, a poor fit requires a correspondingly large $\lambda$. We can measure the difference (or ratio) between the objective change predicted by the quadratic fit ($\hat{f}$) and the real objective change ($f$), by computing $\gamma = (f(w+z) - f(w))/\hat{f}(z)$. This requires one additional evaluation of the objective for $f(w+z)$, but otherwise relies only on previously computed quantities. This makes it a very attractive estimate of the trust region, with $\gamma = 1$ corresponding to a perfect approximation. Following [24], we evaluate $\gamma$ every 5 iterations, decreasing $\lambda$ by a factor of 0.999 when $\gamma > 3/2$, and increasing by the inverse factor when $\gamma < 1/2$. We noted that our algorithm is not very sensitive to the initial $\lambda$. In experiments using batch-normalization (section 4), we simply initialize it to one, otherwise setting it to 10.

## 4 Experiments

### 4.1 Degenerate problems with known solutions

We begin by applying our method to problems of limited complexity, with the goal of exploring the strengths and weaknesses of our approach in an interpretable domain. We perform a comparison with two popular first order solvers — SGD with momentum and Adam [8][2], as well as with more traditional methods such as Levenberg-Marquardt and BFGS [24] (with cubic line-search). The first problem we consider is the search for the minimum of the two-dimensional Rosenbrock test function, which has the useful benefit of enabling us to visualise the trajectories found by each optimiser. Specifically, we use the stochastic variant of this function [26], $\mathcal{R} : \mathbb{R}^2 \to \mathbb{R}$:

$$\mathcal{R}(u,v) = (1-u)^2 + 100\epsilon_i(v - u^2)^2, \qquad (20)$$

where at each evaluation of the function, a noise sample $\epsilon_i$ is drawn from a uniform distribution $\mathcal{U}[\lambda_1, \lambda_2]$ with $\lambda_1, \lambda_2 \in \mathbb{R}$ (we can recover the deterministic Rosenbrock function with $\lambda_1 = \lambda_2 = 1$). To assess robustness to noise, we compare each optimiser on the deterministic formulation and two stochastic variants (with differing noise regimes). We also consider a second problem of interest, recently introduced by Rahimi and Recht [18]. It consists of fitting a deep network with only two linear layers to a dataset where sample inputs $x$ are related to sample outputs $y$ by the relation $y = Ax$, where $A$ is an ill-conditioned matrix (with condition number $\epsilon = 10^5$).

The results are shown in Table 1. We use a grid-search to determine the best hyperparameter settings for both SGD and Adam (details can be found in the supplemental material). We report the number of iterates taken to reach the solution, with a tolerance of $\tau = 10^{-4}$. Statistics are computed over 100 runs of each optimiser. We can observe that first-order methods perform poorly in all cases, and moreover show a very high variance of results. The Levenberg-Marquardt (LM) optimiser generally performs best, although it is impractical for larger-scale problems. Our method delivers comparable performance despite avoiding a costly Hessian inversion. On the other hand, the performance of BFGS, which approximates the Hessian using a rolling buffer of parameter updates, seems to correlate negatively with the level of noise that is present.

---

[2]We also experimented with RMSProp [22], AdaGrad [5] and AdaDelta [27], but found that these methods consistently underperformed Adam and SGD on these "toy" problems.

Table 1: **Optimiser comparison on small degenerate datasets.** For each optimiser, we report the mean±standard deviation of the number of iterates taken to reach the solution. For the stochastic Rosenbrock function, $\mathcal{U}[\lambda_1, \lambda_2]$ denotes noise drawn from $\mathcal{U}[\lambda_1, \lambda_2]$ (see Sec. 4.1 for details)

|  | Deterministic | Rosenbrock $\mathcal{U}[0,1]$ | $\mathcal{U}[0,3]$ | Raihimi & Recht |
|---|---|---|---|---|
| SGD + momentum | $370 \pm 40$ | $846 \pm 217$ | $4069 \pm 565$ | $95 \pm 2$ |
| Adam [8] | $799 \pm 160.5$ | $1290 \pm 476$ | $2750 \pm 257$ | $95 \pm 5$ |
| Levenberg-Marquardt [14] | $16 \pm 4$ | $\mathbf{8 \pm 3}$ | $\mathbf{11 \pm 5}$ | $9 \pm 4$ |
| BFGS [24] | $19 \pm 4$ | $44 \pm 21$ | $63 \pm 29$ | $43 \pm 21$ |
| CURVEBALL (proposed) | $\mathbf{13 \pm 0.5}$ | $12 \pm 1$ | $13 \pm 1$ | $35 \pm 11$ |

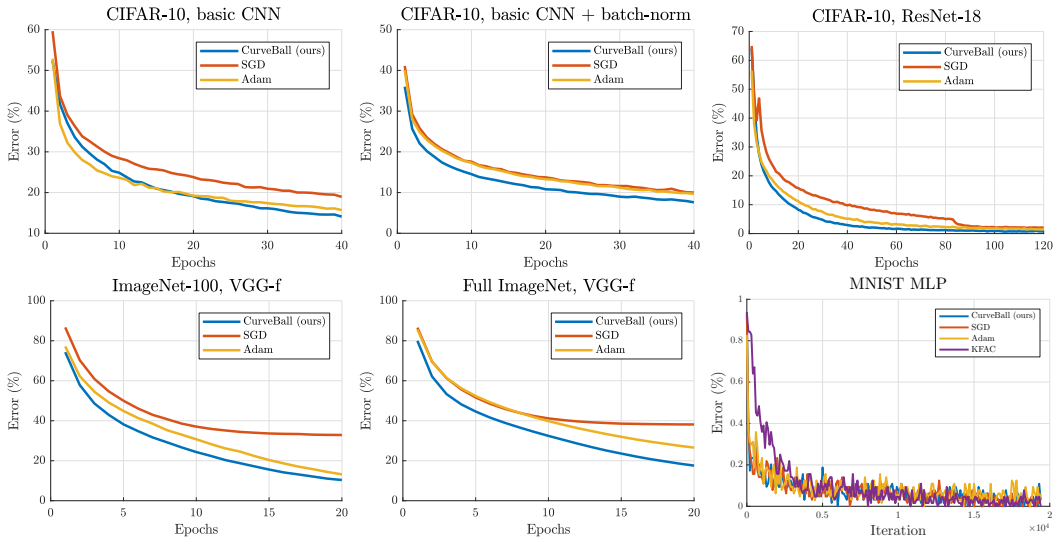

Figure 2: **Comparison with the different optimisers for various datasets and networks**. The evolution of the training error is shown, as it is the quantity being optimised. Our method seems to perform well under a variety of realistic settings, including large-scale datasets (ImageNet), the presence of batch-normalization, and severely over-parameterised models (ResNet).

Fig. 1 shows example trajectories. The slow, oscillating behaviour of first-order methods is noticeable, as well as the impact of noise on the BFGS steps. On the other hand, both our method and LM converge in few iterations.

### 4.2 CIFAR

We now turn to the task of training larger models on more realistic datasets.[3] Second-order methods are typically not used in such scenarios, due to the large number of parameters and stochastic sampling. We start with a basic 5-layer convolutional neural network (CNN).[4] We train this network for 20 epochs on CIFAR-10, with and without batch-normalization (which is known to improve optimisation). To assess optimiser performance on larger models, we also train a much larger

---
[3]Code is available at: https://github.com/jotaf98/curveball
[4]The basic CNN has $5 \times 5$ filters and ReLU activations, and $3 \times 3$ max-pooling layers (with stride 2) after each of the first 3 convolutions. The number of output channels are, respectively, $(32, 32, 64, 64, 10)$.

Table 2: Best error in percentage (training/validation) for different models and optimisation methods. CURVEBALL $\lambda$ denotes use of $\lambda$ rescaling (sec. 3). Numbers in bracket show validation error with additional dropout regularisation (rate 0.3). The first three columns are trained on CIFAR-10, while the fourth is trained on ImageNet-100.

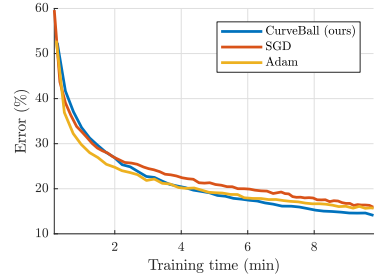

Figure 3: Training error vs. wall clock time (basic CIFAR-10 model).

| Model | Basic | Basic + BN | ResNet-18 | VGG-f |
|---|---|---|---|---|
| CURVEBALL $\lambda$ | **14.1** / 19.9 | **7.6** / 16.3 | **0.7** / 15.3 (13.5) | **10.3** / **33.5** |
| CURVEBALL | 15.3 / **19.3** | 9.4 / **15.8** | 1.3 / 16.1 | 12.7 / 33.8 |
| SGD | 18.9 / 21.1 | 10.0 / 16.1 | 2.1 / **12.8** | 32.9 / 41.7 |
| Adam | 15.7 / 19.7 | 9.6 / 16.1 | 1.4 / 14.0 | 13.2 / 35.9 |

ResNet-18 model [7]. As baselines, we picked SGD (with momentum) and Adam, which we found to outperform the competing first-order optimisers. Their learning rates are chosen from the set $10^{-k}$, $k \in \mathbb{N}$ with a grid search for the basic CNN, while for the ResNet SGD uses the schedule recommended by the authors [7]. We focus on the training error, since it is the quantity being optimised by eq. 1 (validation error is discussed in section 4.6). The results can be observed in fig. 2 (top row). We observe that in each setting, CURVEBALL outperforms its competitors, in a manner that is robust to both normalisation and model type.

### 4.3 ImageNet

To assess the practicality of our method at larger scales, we apply it to the classification task on the large-scale ImageNet dataset. We report results of training on both a medium-scale setting using a subset formed from the images of 100 randomly sampled classes as well as the large-scale setting, by training on the full dataset. Both experiments use the VGG-f architecture and follow the settings described in [3]. The results are depicted in Fig. 2. We see that our method provides compelling performance against popular first order solvers in both cases, and that interestingly, its margin of improvement grows with the scale of the dataset.

### 4.4 Comparison to other second-order methods on MNIST

In order to compare ours with existing second-order methods, we use the public KFAC [12] implementation made available by the authors and run a simple experiment on the MNIST dataset. In this scenario a four layer MLP (with output sizes 128-64-32-10) with hyperbolic tangent activations is trained on this classification task. We closely follow the same protocol as [12] for layer initialisation and data normalisation for all optimisers. We show results in Fig 2 (bottom row, right) with the best learning rate for each method. On this problem our method performs comparably to first order solvers, while KFAC makes less progress until it has stabilised its Fisher matrix estimation.

### 4.5 Wall-clock time

To provide an estimate of the relative efficiency of each model, Fig. 3 shows wall clock time on the basic CIFAR-10 model (without batch norm). Importantly, from a practical perspective, we observe that our method is competitive with first order solvers. Moreover, our prototype implementation includes custom FMAD operations which have not received the same degree of optimisation as RMAD (back-propagation), and could further benefit from careful engineering.

### 4.6 Overfitting and validation error

While the focus of this work is optimisation, it is also of interest to compare the validation errors attained by the trained models – these are reported Table 2. We observe that models trained with the proposed method exhibit better training and validation error on most models, with the exception of ResNet where overfitting plays a more significant role. However, we note that this could be addressed with better regularisation, and we show one such example, by also reporting the validation error with a dropout rate of 0.3 in brackets.

## 5 Related work

While second order methods have proved to be highly effective tools for optimising deterministic functions [24, 14] their application to stochastic optimisation, and in particular to deep neural networks remains an active area of research. A broad spectrum of methods have been recently developed to improve stochastic optimisation by accounting for curvature of the objective function to prevent the loss from getting stuck in 'valleys' [4], while avoiding the cost of storing and inverting the full Hessian. A popular approach has been to construct updates from a buffer of parameter gradients and their first-and-second-order moments at previous iterates (e.g., AdaGrad [5], AdaDelta [27], RMSProp [22] or Adam [8]). These solvers have the benefit of requiring no additional function evaluations beyond traditional minibatch stochastic gradient descent. Typically they set adaptive learning rates by making use of empirical estimates of the curvature with a *diagonal* approximation to the Hessian (*e.g.* [27]) or a rescaled diagonal Gauss-Newton approximation (*e.g.* [5]). While the use of diagonal Hessian decreases the computational cost of these algorithms, their overall efficiency remains limited and in many cases can be matched by a well tuned SGD solver [23].

Second order solvers take a different approach, investing more computation per iteration in the hope of achieving higher quality updates. To achieve this higher quality, they rely on the inversion of the Hessian matrix $H$, or a tractable approximation of this quantity. Perhaps the most popular such approach is to make use of the Gauss-Newton (GN) estimate [11, 13, 2], which uses the Jacobian to form a Positive Semidefinite (PSD) approximation of the Hessian.

Another family of approaches, which have proven effective in the machine learning community for tasks such as classification, introduce second order information with *natural gradients* [1]. In this context, it is common to derive a loss function from a Kullback-Leibler (KL) divergence. The natural gradient makes use of the infinitesimal distance induced by the latter to follow the curvature in the Riemannian manifold equipped with this new distance. In practice the natural gradient method amounts to replacing the Hessian $H$ in the modified gradient formula $H^{-1}\nabla_\theta h$ with the Fisher matrix $F$, which facilitates traversal of the optimal path in the metric space induced by the KL-divergence. Since the seminal work of Amari [1] several authors have studied and implemented variations of the natural gradient idea. The TONGA method [19] relies on the empirical Fisher matrix where the previous expectation over the model predictive distribution is replaced by the sample predictive distribution. The works of [16, 10] established a link between Gauss-Newton methods and the natural gradient. More recently [12] introduced the KFAC optimiser which uses a block diagonal approximation of the Fisher matrix. This was shown to be an efficient stochastic solver in several settings, but it remains a computationally challenging approach for solving large-scale problems with deep neural networks.

Many of the methods discussed above perform an explicit system inversion that can often prove prohibitively expensive [25]. Consequently, a number of works [11, 13, 28] have sought to exploit the cheaper computation of the vector product with $H$ by using automatic differentiation [17, 20] to perform system inversion with conjugate gradients. While these methods have had some success, in practice they can be very sensitive to noise and still often require many steps to solve the system efficiently [21]. Perhaps most closely related to our approach, [15] uses automatic differentiation to compute Hessian-vector products to construct adaptive, per-parameter learning rates.

## 6 Conclusions and future work

In this work, we have proposed a practical second-order solver that has been specifically tailored for deep-learning-scale stochastic optimisation problems. We showed that our optimiser can be applied to a large range of datasets and reach better training error than first order method with the same number of iterations, with essentially no hyper-parameters tuning. In future work, we intend to bring more improvements to the wall-clock time of our method by engineering the FMAD operation to the same standard as back-propagation, and study optimal trust-region strategies to obtain $\lambda$ in closed-form.

**Acknowledgements.** The authors gratefully acknowledge the support of ERC 677195-IDIU, ERC DFR01600, and EPSRC CDT AIMS EP/L015897/1.